\documentclass[conference]{IEEEtran}
\IEEEoverridecommandlockouts
\usepackage{cite}
\usepackage{amsmath,amssymb,amsfonts}
\usepackage{algorithmic}
\usepackage{graphicx}
\usepackage{textcomp}
\usepackage{xcolor}
\usepackage{subcaption}
    
\begin{document}

\title{Lode Encoder:\\AI-constrained co-creativity}

\author{
    \IEEEauthorblockN{Debosmita Bhaumik}
    \IEEEauthorblockA{\textit{Game Innovation Lab} \\
    \textit{New York University}\\
    Brooklyn, USA \\
    db198@nyu.edu}
\and
    \IEEEauthorblockN{Ahmed Khalifa}
    \IEEEauthorblockA{\textit{Game Innovation Lab} \\
    \textit{New York University}\\
    Brooklyn, USA \\
    ahmed@akhalifa.com}
\and
    \IEEEauthorblockN{Julian Togelius}
    \IEEEauthorblockA{\textit{Game Innovation Lab} \\
    \textit{New York University}\\
    Brooklyn, USA \\
    julian@togelius.com}
}

\IEEEoverridecommandlockouts
\IEEEpubid{\makebox[\columnwidth]{978-1-6654-3886-5/21/\$31.00 ©2021 IEEE \hfill} \hspace{\columnsep}\makebox[\columnwidth]{ }}
\maketitle
\IEEEpubidadjcol

\begin{abstract}
We present Lode Encoder, a gamified mixed-initiative level creation system for the classic platform-puzzle game Lode Runner. The system is built around several autoencoders which are trained on sets of Lode Runner levels. When fed with the user's design, each autoencoder produces a version of that design which is closer in style to the levels that it was trained on. The Lode Encoder interface allows the user to build and edit levels through ``painting'' from the suggestions provided by the autoencoders. Crucially, in order to encourage designers to explore new possibilities, the system does \emph{not} include more traditional editing tools. We report on the system design and training procedure, as well as on the evolution of the system itself and user tests.
\end{abstract}


\section{Introduction}

It has been said that constraints are good for creativity~\cite{acar2019creativity}. Some of the greatest art in the world came about through adhering to constraints, either self-imposed or imposed by others. Beginners and professional artists alike often find themselves becoming more productive (e.g. escaping writers' block) or gaining new perspectives and methods through carrying out formal exercises. One reason that constraints can facilitate the creative process may be that they can steer creators away from familiar solutions and workflows.

AI-assisted tools for co-creativity typically seek to assist a human designer through providing suggestions and feedback~\cite{yannakakis2014mixed,smith2010tanagra,guzdial2019friend,liapis2013sentient}. The basic ideas is to use various AI methods to outsource part of our cognitive labor, such as playtesting a level, fixing color balance, or suggesting where to place a character. An implicit assumption is that the tool is attempting to support the human creator in making the artifact the human wants to create. Michelangelo famously said ``The sculpture is already complete within the marble block, before I start my work'' and that he just had to ``chisel away the superfluous material''. The guiding philosophy behind many of these AI-assisted co-creation tools seem to be to help discover the sculpture that is already in the artist's mind, and suggest helpful parts to chisel away. Some of these tools explicitly try to model the style of the user or the end result they wanted to create.

But what would happen if we allowed the AI to assist (or ``assist'') the human creator by setting constraints, or making suggestions that are orthogonal to the direction of the user? This paper explores this question by describing \emph{Lode Encoder}\footnote{http://www.akhalifa.com/lodeencoder/}, a system designed to help but also constrain the designer. The target domain for the system is levels for the classic platformer \emph{Lode Runner}, a game which emphasizes real-time puzzle-like gameplay on medium-sized 2D levels. Lode Encoder allows you to create/edit these levels, but not freely; you must compose your levels based on parts of suggestions generated by neural networks. More specifically, these neural networks are autoencoders (thus the name of the system) trained on a corpus of existing Lode Runner levels. The autoencoders take the level, the user is currently editing, and try to make it more like various levels they have been trained on. As the user composes their level, they can ask for more suggestions from the networks, but only a limited number of times, to discourage asking for suggestions until they just know what they want.

This paper first describes the game itself, and related attempts to learn level generators for it and similar games. It then describes the autoencoders, the dataset they were trained on, and data augmentation procedures. We also comment on the somewhat complicated genesis of the system. After describing the Lode Encoder system, we report on the results of an open-ended user study based on recruiting users on Twitter.

\section{Lode Runner}

\begin{figure}
    \centering
    \includegraphics[width=.7\linewidth]{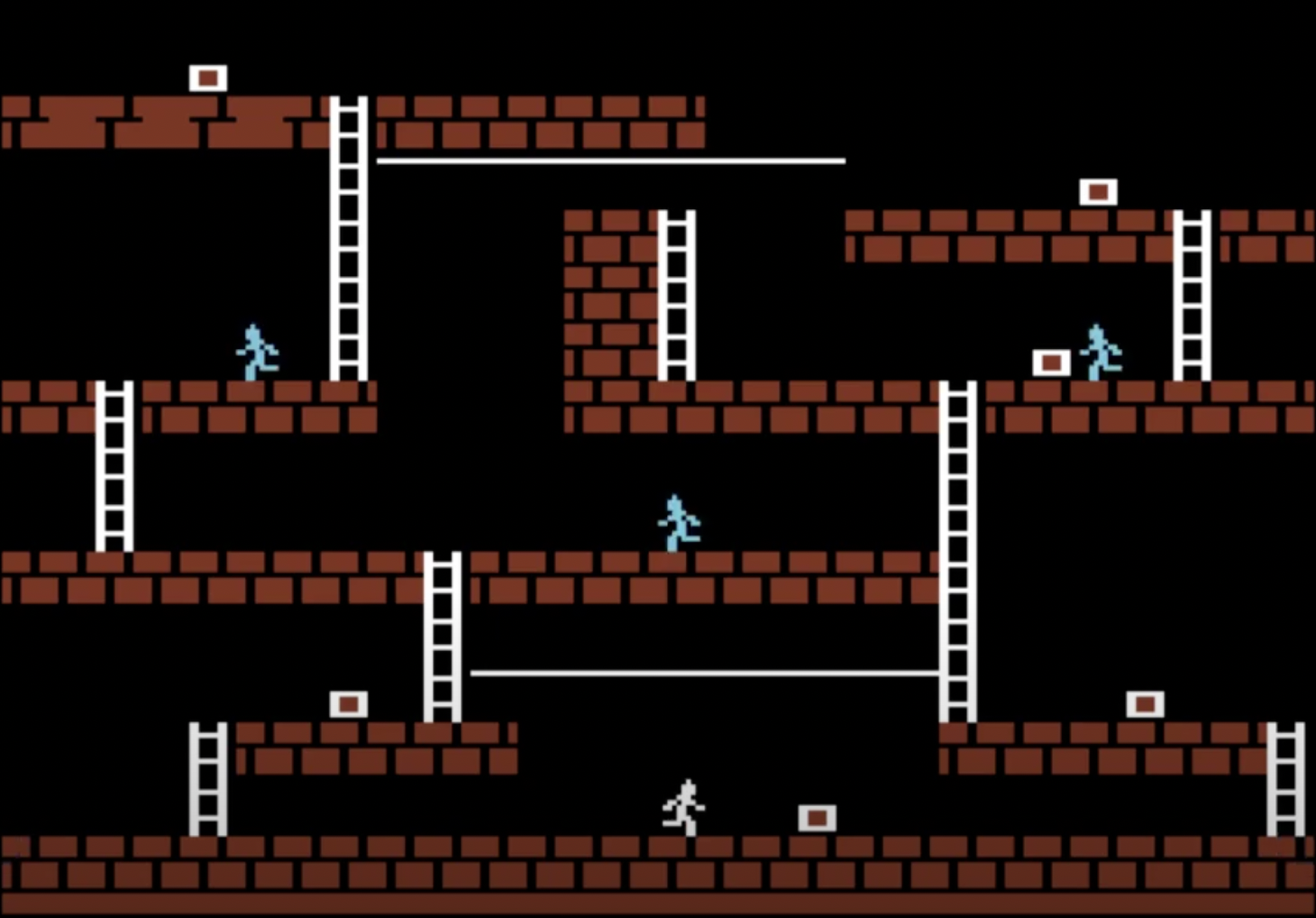}
    \caption{A classic Lode Runner level from Commadore 64 release}
    \label{fig:LodeRunnerExample}
\end{figure}

Lode Runner is a 2D platformer-puzzle game that was first published by Broderbund in 1983 on multiple systems (Apple II, Atari 8-bit, Commodore 64, VIC-20, and IBM PC). The goal of the game is to collect all the gold nuggets in the level while avoiding the enemies that try to catch the player character. The player can traverse levels by walking on platforms, climbing up ladders, moving using ropes, and falling from edges and ropes but, unlike most other platform games, cannot jump. Additionally, the player can dig hole either to their left or right to make path or to trap enemies. These holes are temporary, over time tiles will regenerate to fill the holes. Figure~\ref{fig:LodeRunnerExample} show an example level from the Commodore 64 release.

Due to Lode Runner's success~\cite{ferrell1987commodore}, the game was ported/remaked to a lot of consoles including newer ones such as Nintendo Switch\footnote{https://www.nintendo.com/games/detail/lode-runner-legacy-switch/}. Although of that wide spread, the game didn't get much attention in the AI in Games research community compared to other platformer games such as Super Mario Bros~\cite{pearson2015mario}. Perhaps because of the puzzle-heavy nature of Lode Runner, and there not being a game-playing competition or benchmark for the game, it has been seen as less attractive to work on.

The few papers that target Lode Runner focus on machine learning-based level generation for the game. 
Snodgrass and Ontan{\'o}n~\cite{snodgrass2016learning} applied Markov model to generate levels for Lode Runner, Super Mario Bros and Kid Icarus. Thakkar et al.~\cite{thakkar2019} used vanilla autoencoder to learn latent space of Lode Runner levels and then used latent space evolution~\cite{bontrager2018deep} to generate new Lode Runner levels. Snodgrass and Sarkar~\cite{snodgrass2020multi} trained a Conditional Variational Autoencoder on levels of 7 different games (including Lode Runner) and used it with example-driven binary space partitioning algorithm to generate/blend new levels across these games. Steckel et al.~\cite{steckel2021illuminating} combined the efficiency of GANs with MAP-Elites to generate new levels for Lode Runner. 











\begin{figure*}[ht]
    \centering
    \includegraphics[width=.6\linewidth]{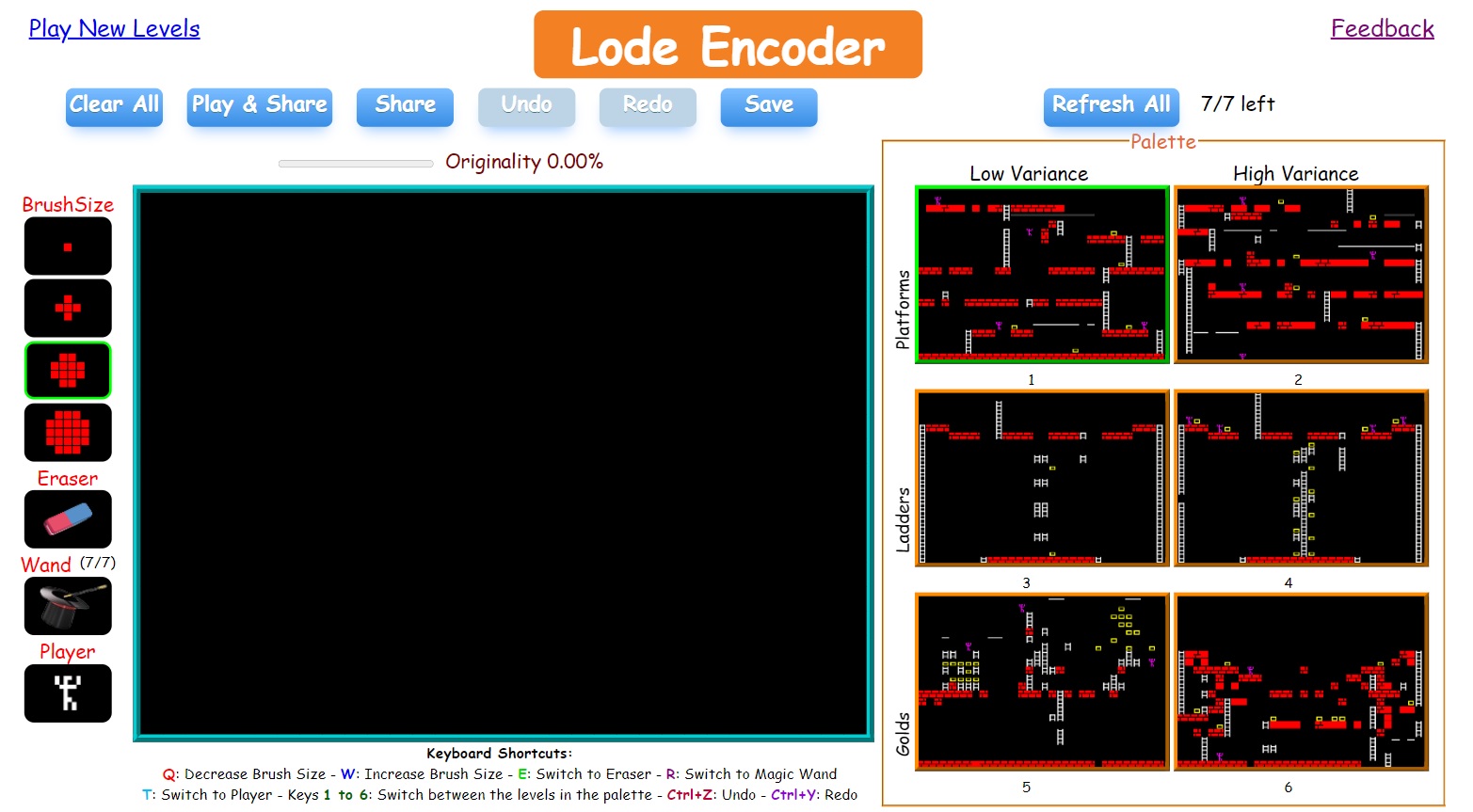}
    \caption{A screenshot of Lode Encoder}
    \label{fig:LodeEncoder}
\end{figure*}

\section{Mixed-initiative game design tools}

Lode Encoder is an AI-constrained mixed-initiative level design tool~\cite{yannakakis2014mixed,liapis2016mixed} that helps user to build Lode Runner levels. The important feature of this tool is the user is constrained by the AI. The system does not give direct control to the user for building levels, instead it provides suggestions generated by the trained Autoencoders to help the user. 

Sarkar and Cooper~\cite{sarkar2020towards} investigated the different ways to use Machine Learning to enable the users to generate, modify, and/or blend levels from a single or multiple games. Based on their work and a number of other AI-assisted mixed-initiative tools in the literature. We can divide the tools according to the role of the AI agents in the system:
\begin{itemize}
    \item \textbf{a supporting role}: the agent provides a group of suggestions to the user where the user can accept it or ask for new ones. For example, Sentient Sketchbook~\cite{liapis2013sentient} is a mixed-initiative tool that supports creating 2D strategy maps for StarCraft. The tool provides the user with multiple suggestions that can improve different factors in the current level such as balance, safe area, etc. Alvarez et al.~\cite{alvarez2019empowering} followed the same philosophy and provided a similar tool to generate dungeons for an dungeon crawler game. 
    \item \textbf{a friend:} modify the content directly without asking for the user permission. For example: Tanagra~\cite{smith2010tanagra} is 2D mixed-initative tool that creates 2D platformer levels. The user can add some constraints in these levels by fixing certain areas and the system generate the whole level while respecting the user constraints. Similarly, Guzdial et al.~\cite{guzdial2019friend} developed a game design tool where the user can build Super Mario Bros levels with the help of a trained AI agent. The user and the AI agent take turns in creating the level and the agent adapts to the user style by updating its models using active learning methods.
    \item \textbf{a creative force:} the agent creates the whole content and the user can only direct the agent using some auxiliary inputs. For example, Picbreeder~\cite{secretan2008picbreeder} and Artbreeder~\cite{simon2015artbreeder} evolve pictures with the help of the user by allowing the user to be the critic. The system allow the user to select levels that they like then the system take these levels and blend them together to allow for new images inspired by the user selection. Similary, Schrum et al.~\cite{schrum2020interactive} uses similar evolution technique to generate levels for Super Mario Bros (Nintendo, 1985) and The Legend of Zelda (Nintendo, 1986). The main difference is that they evolve the latent variables of a trained Generative Adversarial Network (GAN). Schubert et al.~\cite{schubert2021toad} provided a UI that allow the user to generate Super Mario Bros (Nintendo, 1985) levels using their TOAD-GAN. TOAD-GAN is a new proposed GAN that can generate new levels from a single data point. The tool provided with the system gives the user minimal control such as which trained GAN to use, the similarity to the training data set, etc.
\end{itemize}
Lode Encoder, the subject of this paper, doesn't fit easily in any of these categories. It blurs the line between the agent in a supporting role as the user still have full creative control on what to place and where to place, and the agent as a creative force as the agent constraints all these values and the user just control the generated suggestion using an auxiliary input (the current level).

\section{Autoencoder and Training}
Autoencoders~\cite{bourlard1988auto,kramer1991nonlinear} are unsupervised neural networks that learn to compress data in lower dimension. Autoencoders are made of two parts, an encoder which compresses the input data into lower dimension (called latent space) and a decoder that reconstructs the data from the latent space. 
Autoencoders have previously used too generate levels, repair levels, and recognize design style~\cite{jain2016autoencoders}.

While vanilla autoencoders learn to map the input with a single latent space, Variational Autoencoders(VAE)~\cite{an2015variational} learn the probability distribution of latent space for a given input set, which enables random sampling and interpolation of output. This can also enable a better representation for the latent space which can in tern allow an easier way to search the latent space for contents.

In this work, we use VAEs, as they in preliminary tests showed better results than vanilla autoencoders. Our network architecture is based on Sarkar et al. work~\cite{sarkar2020conditional,sarkar2021generating}. Both our encoder and decoder are made of 4 fully-connected layers with batch normalization after each layer. The encoder uses ReLU activation for each layer, whereas the decoder uses ReLU for first three layers and Softmax in the final layer.

\subsection{Dataset}

We used all the 150 classic Lode Runner levels for our experiments. The level data is taken from Video Games Level Corpus (VGLC)\cite{summerville2016vglc}. These levels are consisted of 22x32 tiles, each tile belongs to one of the tile type from the tileset: solid, breakable, enemy, gold, ladder, rope, and empty. We encoded the data using one-hot encoding. This number of levels is relatively large compared to other games for which PCGML\cite{summerville2018procedural} have been attempted, but small by the standards of machine learning in general. Machine learning algorithms usually achieve better results when there is an abundance of training data which is feasible in fields like face generation, text generation, etc (Imagenet has more than 14 million images). This is not the case in AI in Games research as game data is either small (Super Mario Bros has dozen of levels) or publicly not available (Super Mario Maker levels can't be accessed outside of the game). 

One of the most common solutions to that problem is to increase the dataset size by augmenting the input data~\cite{mikolajczyk2018data,shorten2019survey}. We applied padding to each of the levels using solid tiles. Each level of the dataset was padded with 10 columns. Varying the number of columns padded to left and right we create 11 padded levels from each original level. We started with no padded column on left and 10 padded columns on right of a level then increased the number of padded columns on left and decreased the numbers on right. We repeated this process until left had 10 padded columns and no padded columns on right for each original level. Then each of these level can be reflected across the x-axis to double of the size of the levels. The total amount of levels after augmentation is 3300 instead of 150.

\begin{figure*}[ht]
    \centering
    \includegraphics[width=.6\linewidth]{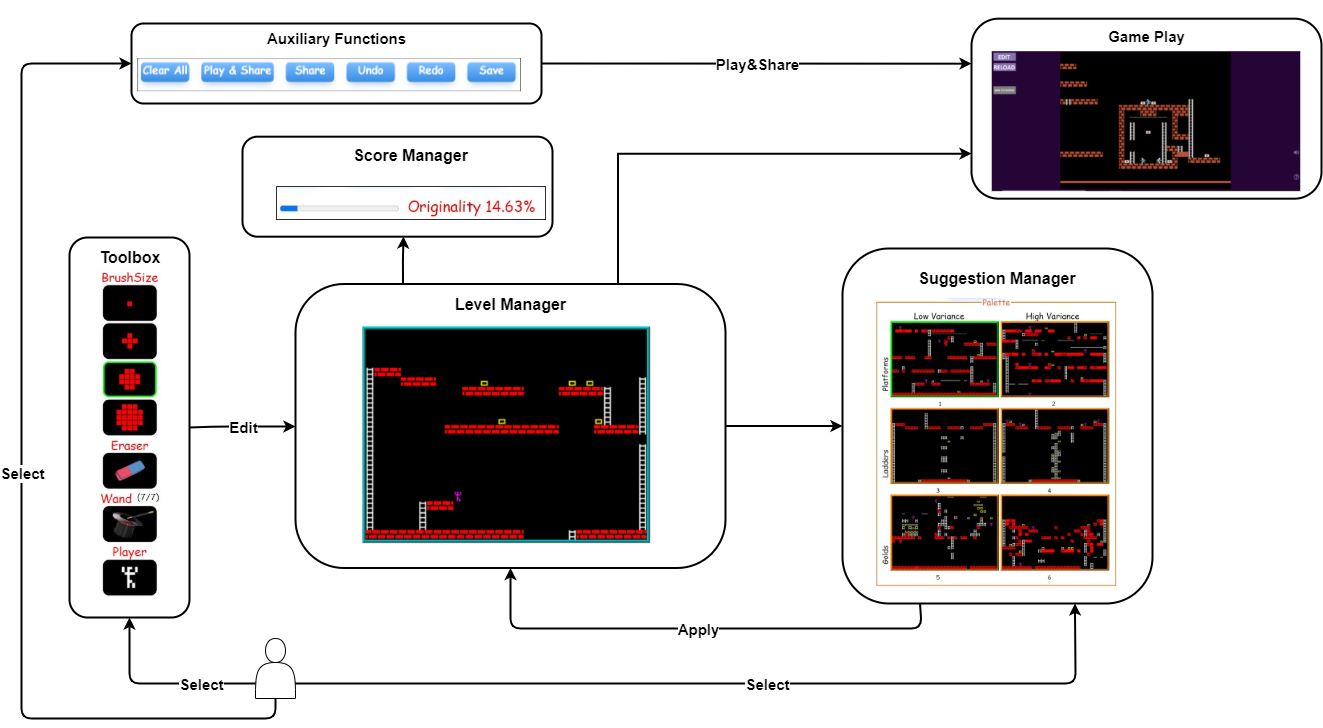}
    \caption{Lode Encoder system architecture}
    \label{fig:LodeEncoderSystemDesign}
\end{figure*}

\subsection{Training}
For training, we divided the dataset into three sets of the original levels. These sets were created based on the features of the levels: levels with lots of gold, levels with platforms, and levels with long ladders or multiple ladders. This split of dataset was made manually. Each set contains 50 levels, which after data augmentation as described above become 1100 training instances. We trained three different VAE models on each of these sets and an additional VAE on the whole dataset. For the rest of the paper we refer to the VAEs as VAE-Gold, VAE-Platform, VAE-Ladder, and VAE-All.

Each VAE was trained for 10,000 epochs using the Adam optimizer. Default learning rate was 0.001, learning rate decayed every 2500 epochs by 0.01. We used Categorical-Cross Entropy to calculate the loss of the model. In the initial experiments we tried different sizes of the latent space 32, 64 and 128. Based on the results, we finally settled on the latent size of 128 as it provided the best result.

\section{Lode Encoder}

Figure~\ref{fig:LodeEncoder} shows a screenshot of our Lode Encoder tool. The tool provides different size brushes (on the left) that the user can use to paint from the suggested levels (on the right) in the editor canvas. The system encourages users to mix levels by providing an originality score (above the canvas). This score reflects how different and new the created level from the original training levels. Finally, the system limits the amount of times the user can get new suggestion (above the suggestions). This restriction was added to push users to make the best use of the current suggestions. Beside level creation, the system facilitates testing and sharing levels online over Twitter (above the originality score).


Figure~\ref{fig:LodeEncoderSystemDesign} shows the overall system architecture of Lode Encoder. The system consists of six different modules:
\begin{itemize}
    \item \textbf{Level Manager:} maintains the current level and enables the user to manipulate it using the toolbox and auxiliary functions.
    \item \textbf{Toolbox:} enables the user to select a tool for editing the current level.
    \item \textbf{Suggestion Manager:} generates and display the AI generated suggestion.
    \item \textbf{Auxiliary Functions}: provides the user with extra functionalities that manipulates the flow of the tool.
    \item \textbf{Game Player:} allows the user to play and share the current game level.
    \item \textbf{Score Manager:} generates and shows the current originality score.
\end{itemize}
In the following subsections, we are going to discuss each of these modules in details.


\subsection{Level Manager}
Level Manager maintains the current level state during the level creation process. It is the central piece of flow in the whole system. The level manager send the current level to Score Manager to calculate the originality score, Suggestion Manager to generate new suggestions, and Game Player to test and share the current level. The user can only manipulates the current level using the provided tools from the toolbox.


\subsection{Toolbox}
Toolbox provides the main tools to manipulate the current level in the Level Manager. It includes these four tools:
\begin{itemize}
    \item \textbf{Brush Tool}: enables the user to apply segments of suggested levels to the current level. The brush applies the corresponding tiles from the selected suggested level to the current level. The toolbox provides four different brush sizes to provide the user with fine control on the granularity of their painting.
    \item \textbf{Eraser Tool}: allows the user to erase certain tiles in the current level. The eraser provides the users to fine control on removing some of the painted tiles from the suggestions.
    \item \textbf{Wand Tool}: adds the majority voting value to the selected tile. This tool was added in a later iteration (check section~\ref{sec:SystemEvolution} for more details) of the tool to help users to fill small broken areas in their design. This is the only tool that is limited in its usage to a maximum of 7 tiles. This limit prevents the user from over using the tool to build their level and pushes them towards mixing the suggested levels.
    \item \textbf{Player Tool}: provides the user with the ability to choose the player starting position in the level. We provided this tool because the suggestions does not gives any player as the models were not trained on player tile.
\end{itemize}

\subsection{Suggestion Manager}
Suggestion Manager is responsible for providing suggestions from the current level. It uses three different models: VAE-Platform, VAE-Ladder, and VAE-Gold. Each model generates two suggestions based on the current level, one is with low variance to the current level other is a high variance to the current level. 
Current level is feed to the encoder of the VAE and latent vector is obtained then with a small uniform noise (-0.005 $<=$ noise $<$ 0.005) is added to the latent vector then it is fed to the decoder to get the low variant suggestion. While for the high variant suggestion comparatively big uniform noise (-0.5 $<$ noise $<$ 0.5) is added to the latent vector and the loop goes for 10 times. This loops is added to help the suggested level to look a lot different from the previous suggestion if the user didn't change anything in the current level.

All the suggestions are displayed in a grid of 3x2 where each row comes from a different model (VAE-PLatform, VAE-Ladder and VAE-Gold respectively) and each column suggest different variance (low variance and high variance respectively). The Suggestion Manager allows the user to select from the suggested levels and use them with the brush tool to make their own level. User can ask for new suggestions but for limited number of times (maximum of 7). New suggestions will be generated based on the current level in the Level Manager. The limited number of refreshes pushes the user to be creative in using the current suggestion as they can't build any level they want.

\subsection{Auxiliary Functions}
Auxiliary Functions provides the user with additional functionality that doesn't affect the current level but makes the user's life easier. The Auxiliary Functions contains:
\begin{itemize}
    \item \textbf{Clear All Function}: help the user to clear everything and restart the whole system instead of refreshing the website, every time the user want to do that.
    \item \textbf{Play \& Share Function}: passes the current level from the Level Manager to Game Player so the user can test their level and share it on Twitter with other people to try.
    \item \textbf{Undo \& Redo Function}: saves all the edits the user do in the tool and allow them to undo/redo any accidents or mistakes they have done during creating a new level.
    \item \textbf{Save Function}: generates a URL that contains the current level so the user can share it with a friend or continue editing it later in time.
\end{itemize}

\subsection{Game Player}
Game Player enables the user to play the current level. This helps users to test their current level and see if they are playable or not. Once the user wins the level, the level can be shared on Twitter, or elsewhare using a link. This module was not built from scratch, we modified an open source HTML5 implementation of Lode Runner by Simon Hung\footnote{https://github.com/SimonHung/LodeRunner\_TotalRecall}. Currently we do not have any automated agent to determine the playability.

\subsection{Score Manager}
The Score Manager finds the originality value of the current level. Any change in the current level updates the score. Higher originality value indicates less similarity between training levels and current level. The orginality score is calculated using the VAE-All model. We pass the current level through the VAE-All model and then measure the hamming distance between the reconstructed level and the input level. This hamming distance measures the number of different tiles between both of them. We normalize that value by dividing by the level area and then show it to the user as the final score. To make the user notice the score more and force them to make a good level, we made sure the originality score flashes red if it is less than 25\%.

\begin{table}
    \centering
    \begin{tabular}{|l|r|}
        \hline
        Name  &  Statistics\\
        \hline
        Number of users & 100\\
        Number of sessions &  132\\
        Number of tested levels & 84\\
        Number of playable levels & 24\\
        Number of Feedback & 7\\
        \hline
    \end{tabular}
    \caption{Statistics on using the tool over the course of 18 days}
    \label{tab:stats}
\end{table}

\section{System Evolution}\label{sec:SystemEvolution}

The Lode Encoder tool came out of a project seeking to train autoencoders to generate content. We wanted to see if we could make autoencoders, a self-supervised learning method, generate levels as well as Generative Adversarial Networks (GANs), another self-supervised learning method. If this worked, we would use the basic idea of Latent Variable Evolution, searching the latent inputs of GANs for vectors that lead to good content, but using the bottleneck layer of the autoencoder. In other words, the initial focus was on autonomous rather than interactive content generation. 

Alas, things went less than perfectly well. Although we tried every trick we could think of, the autoencoders generally produced rather lousy levels, that rarely passed our automated playability test. They seemed unable to model the long-distance relationships between level elements that are so important for a game like this; the playability of a level depends on that every gold nugget is reachable from the starting point, something that can be hard to ascertain. It is worth pointing out that the spatial dependencies that must be understood to make a playable level are much more complex than those of a typical side-scrolling Super Mario Bros level, the latter being a target of much PCGML research recently.
The size of the level in tiles is also substantially larger compared to a single screen (the standard unit of generation) in Super Mario Bros. This may explain the partial success.

One thing we discovered was that we could train autoencoders to generate almost-perfect levels, but only if we overfit them badly on the data. In other words, regardless of the input, the overfitted networks would always output something close to a level in the training set; furthermore, if we pass the output of the network back into it a few times, it will converge on something almost identical to a training level. This of course limits the usefulness of the networks as autonomous content generators. But it gave us another idea: What if we used the overfitted networks as a kind of repair function, which would take what a user created and move it in the direction of something the network knows to be a good level?

With this in mind we built our first version of Lode Encoder as a mixed initiative tool. This first version included traditional editing functionality--``painting'' the different tile types directly onto the level canvas--as well as being able to paint from the suggested levels. Informal testing with lab members and friends showed that users mostly did not use the suggestions beyond the initial ideation stage. That is, most users started off their design process by painting from one of the suggested levels, but then they continued using solely the traditional editing tools. When asked why, users would explain that the suggestions were either not good or not what they wanted. Our interpretation of this is that the AI-assistance was not used to its full potential. We wanted the suggestions to help users break away from their initial design ideas, and design around constraints posed by the AI so as to discover new design strategies. We thought that this could be accomplished by incorporating a more gamelike interaction.

\begin{figure}
    \centering
    \includegraphics[width=.7\linewidth]{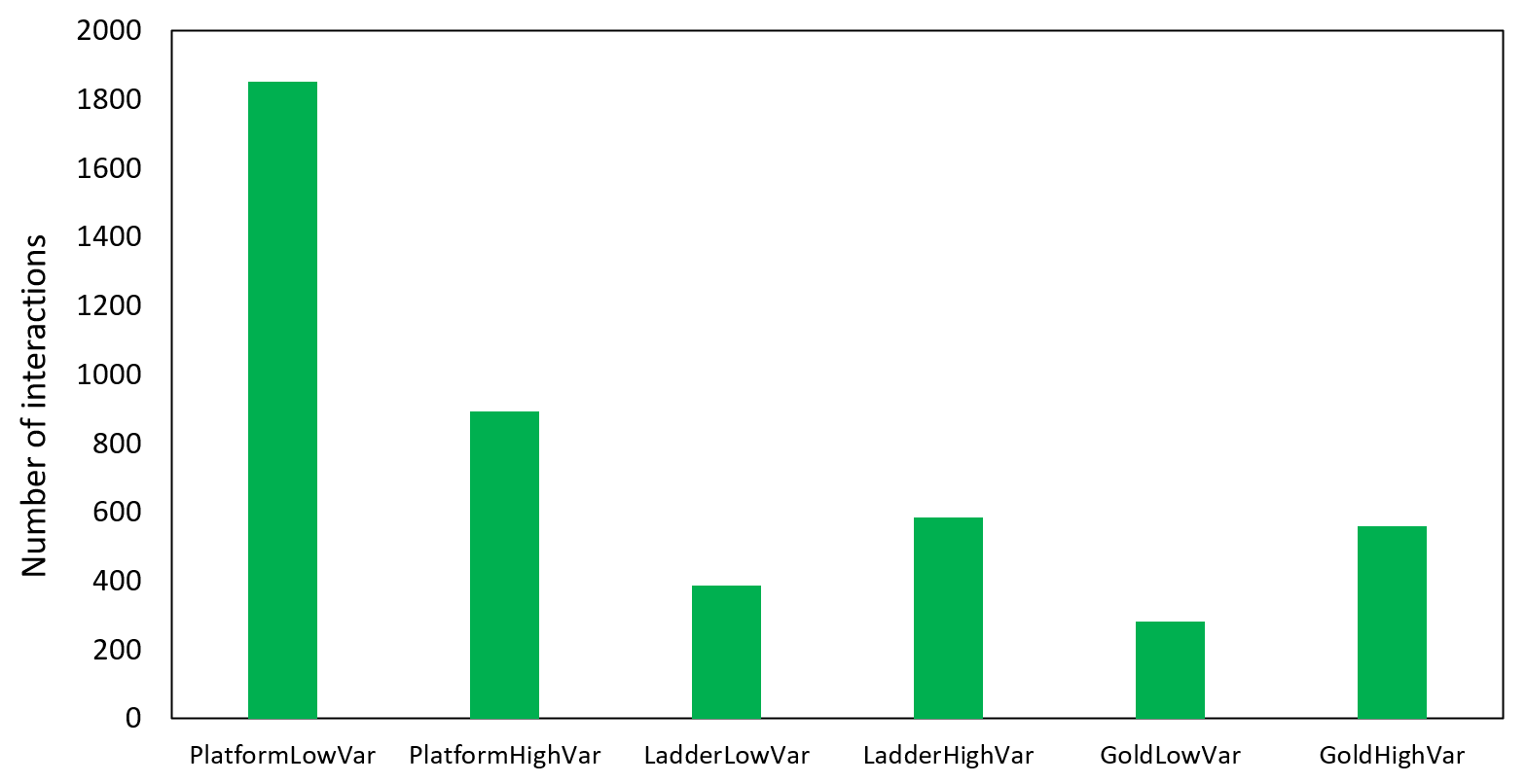}
    \caption{Number of interactions that each suggestion got over all the sessions}
    \label{fig:modelUsed}
\end{figure}

\begin{figure}
    \centering
    \includegraphics[width=.7\linewidth]{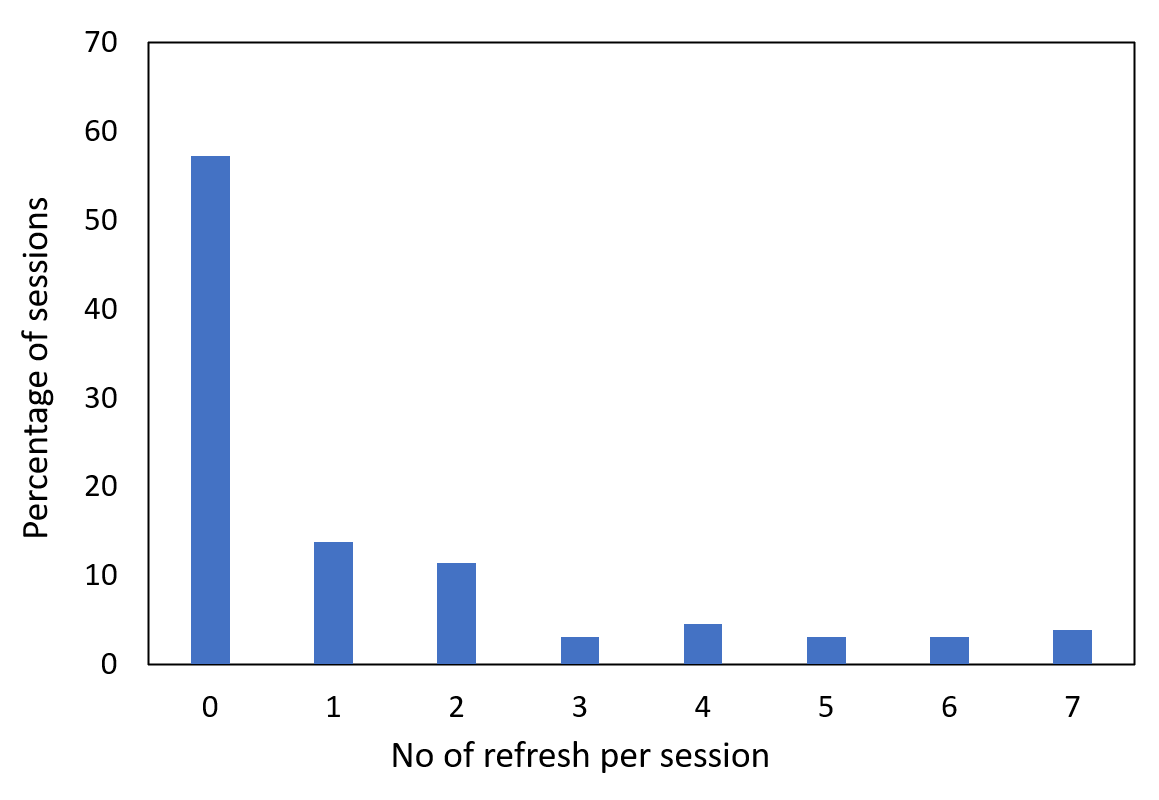}
    \caption{Histogram about the number of refreshes used in every session}
    \label{fig:refreshes}
\end{figure}

We therefore took the radical step of removing the possibility for traditional map editing, i.e. selecting tiles freely and adding them to the level. We also limited the number of times the user could request new suggestions, to avoid that users would refresh suggestions until they found exactly what they wanted. The idea is that the users would have to work with what they were given, and find ways to design around limitations.

This version of the system went through the most testing, and was advertised widely by the authors and others on social networks. While Lode Encoder was generally well received, and many found the interaction fun and interesting, a number of users reported that they found making a playable level a bit of a chore. A typical problem was finding the few missing tiles of ladder or brick that would be needed to make a particular design playable. This was not what we intended with our system; we wanted the user to have to design around limitations in a playful way, but we did not want to make it hard to design a playable level.

To alleviate this problem, we introduced the wand. As described above, the wand simply reproduces the most common tile among its neighbors. This effectively ``repairs'' broken ladders or platforms, and was generally perceived by users as working well.

\section{System Evaluation}

\begin{figure}
    \centering
    \includegraphics[width=.6\linewidth]{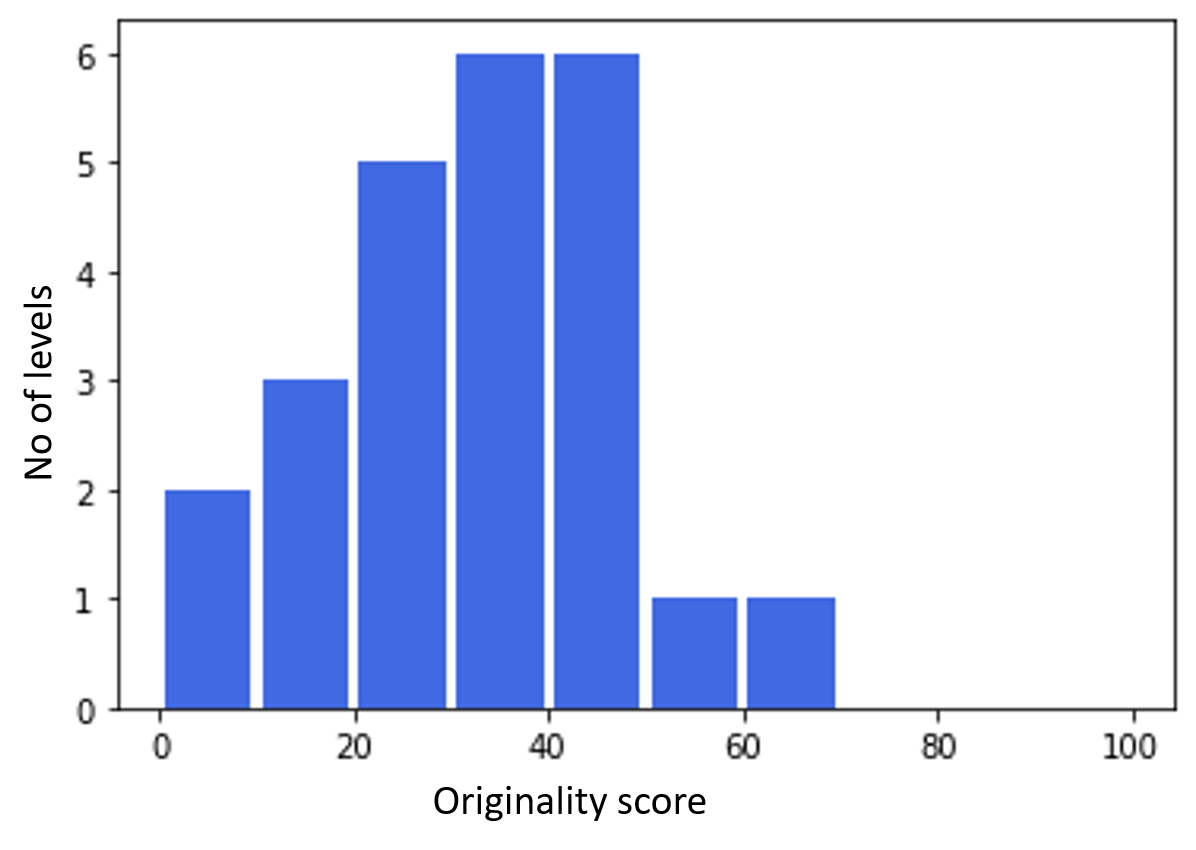}
    \caption{Originality of user made levels}
    \label{fig:originality}
\end{figure}

\begin{figure*}
    \centering
    \captionsetup{justification=centering}
    \begin{subfigure}[t]{\linewidth}
        \centering
        \includegraphics[width=.24\linewidth]{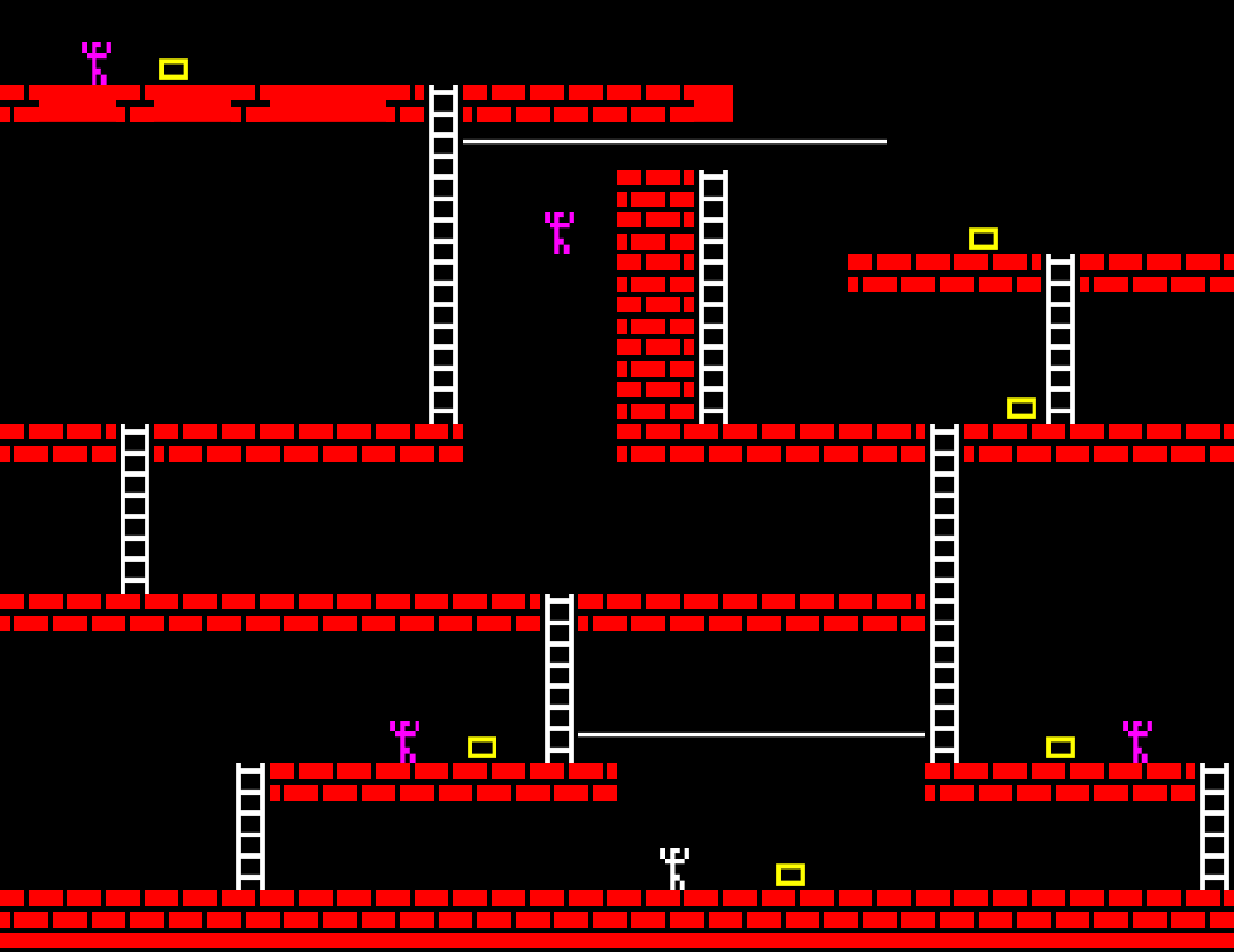}
        \includegraphics[width=.24\linewidth]{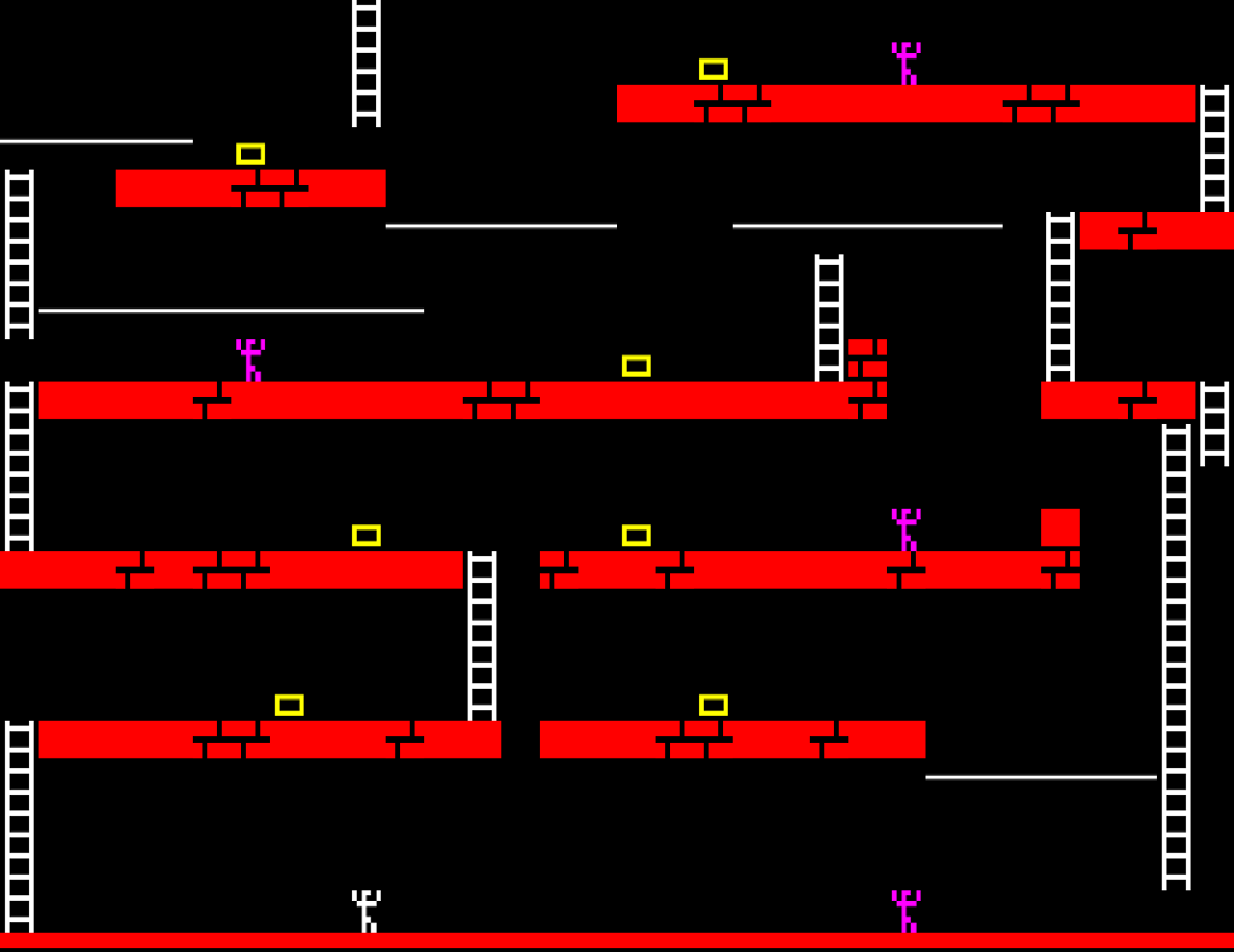}
        \includegraphics[width=.24\linewidth]{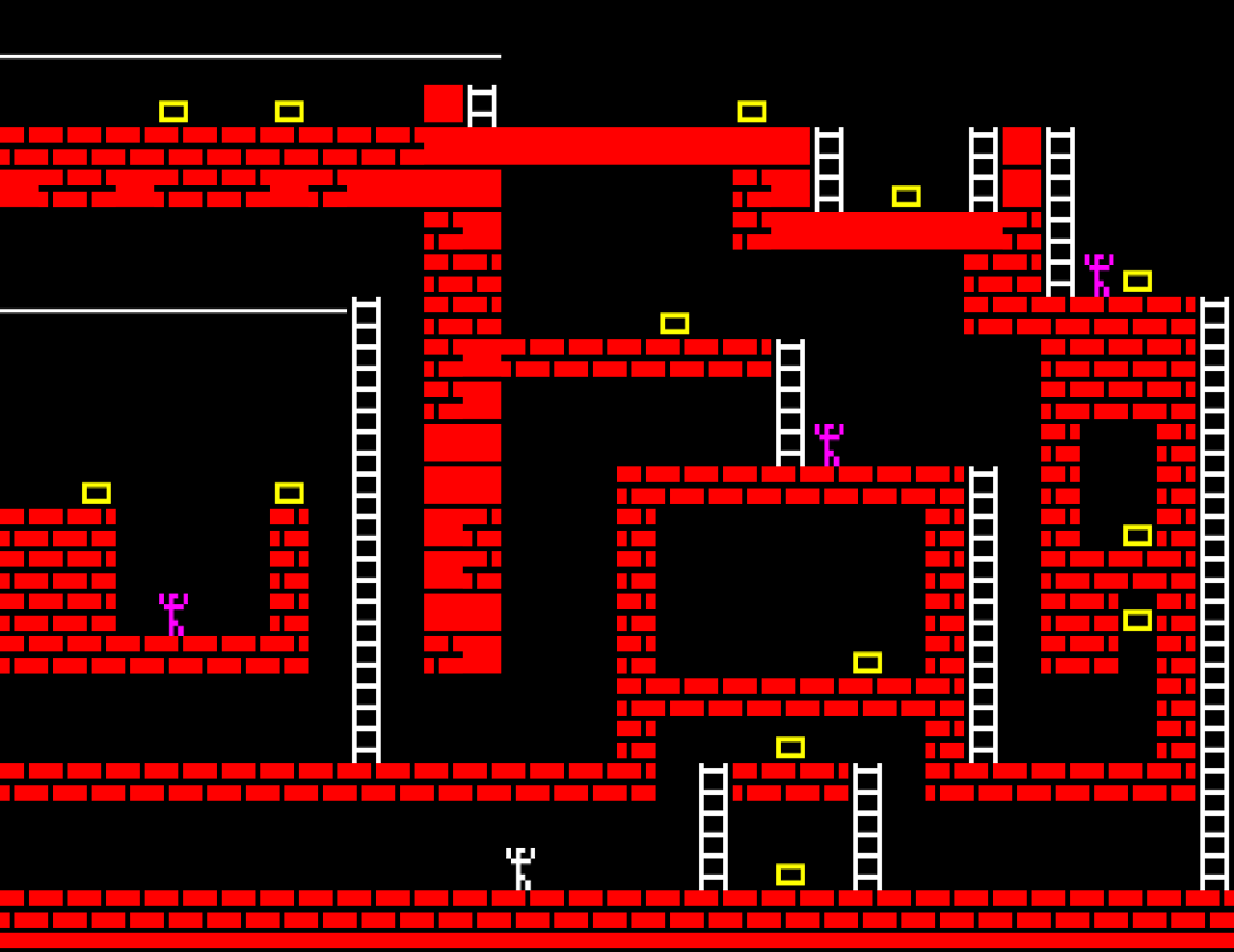}
        \includegraphics[width=.24\linewidth]{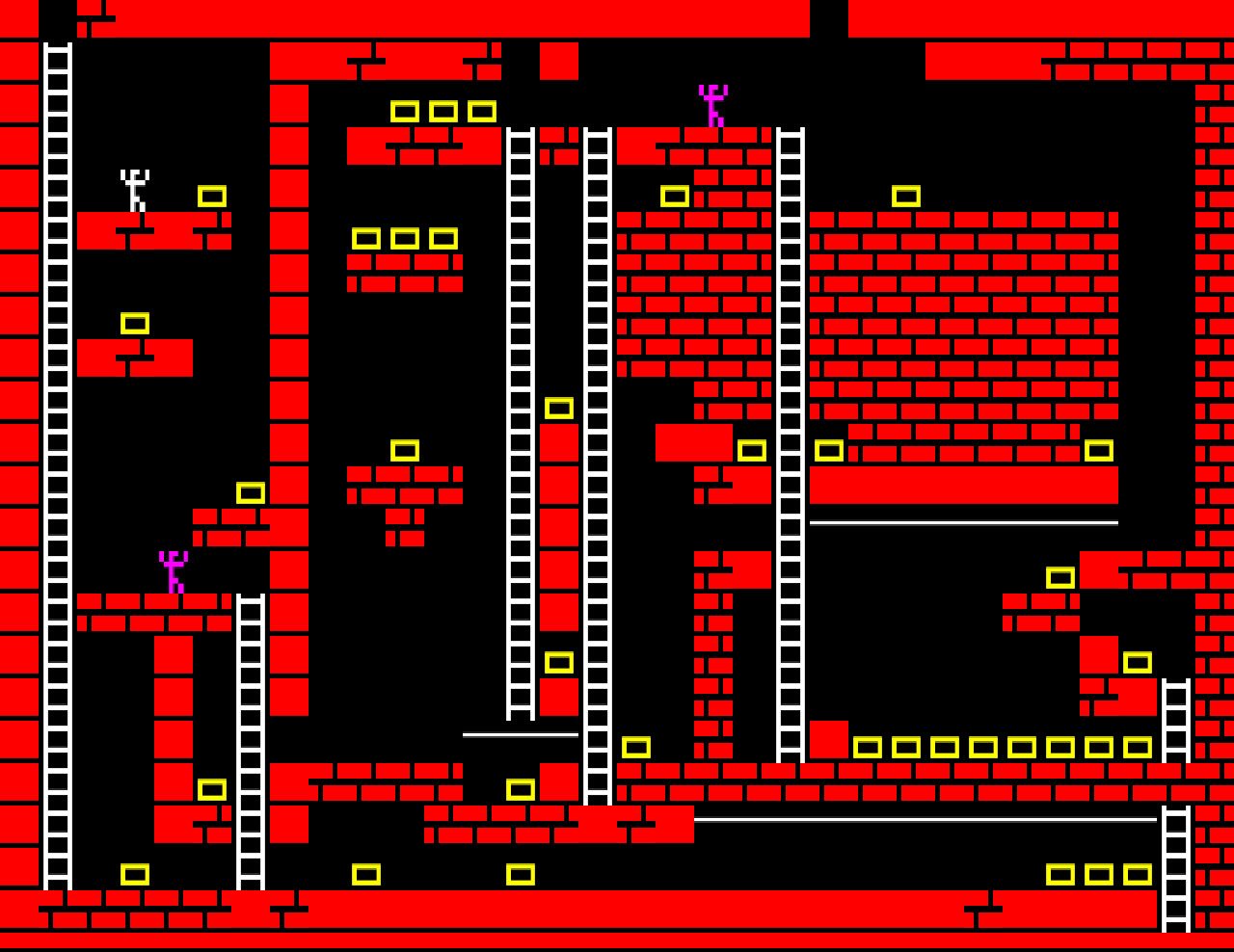}
        \caption{Levels from training dataset.}
        \label{fig:original}
    \end{subfigure}\vspace{.15cm}
    \begin{subfigure}[t]{\linewidth}
        \centering
        \includegraphics[width=.24\linewidth]{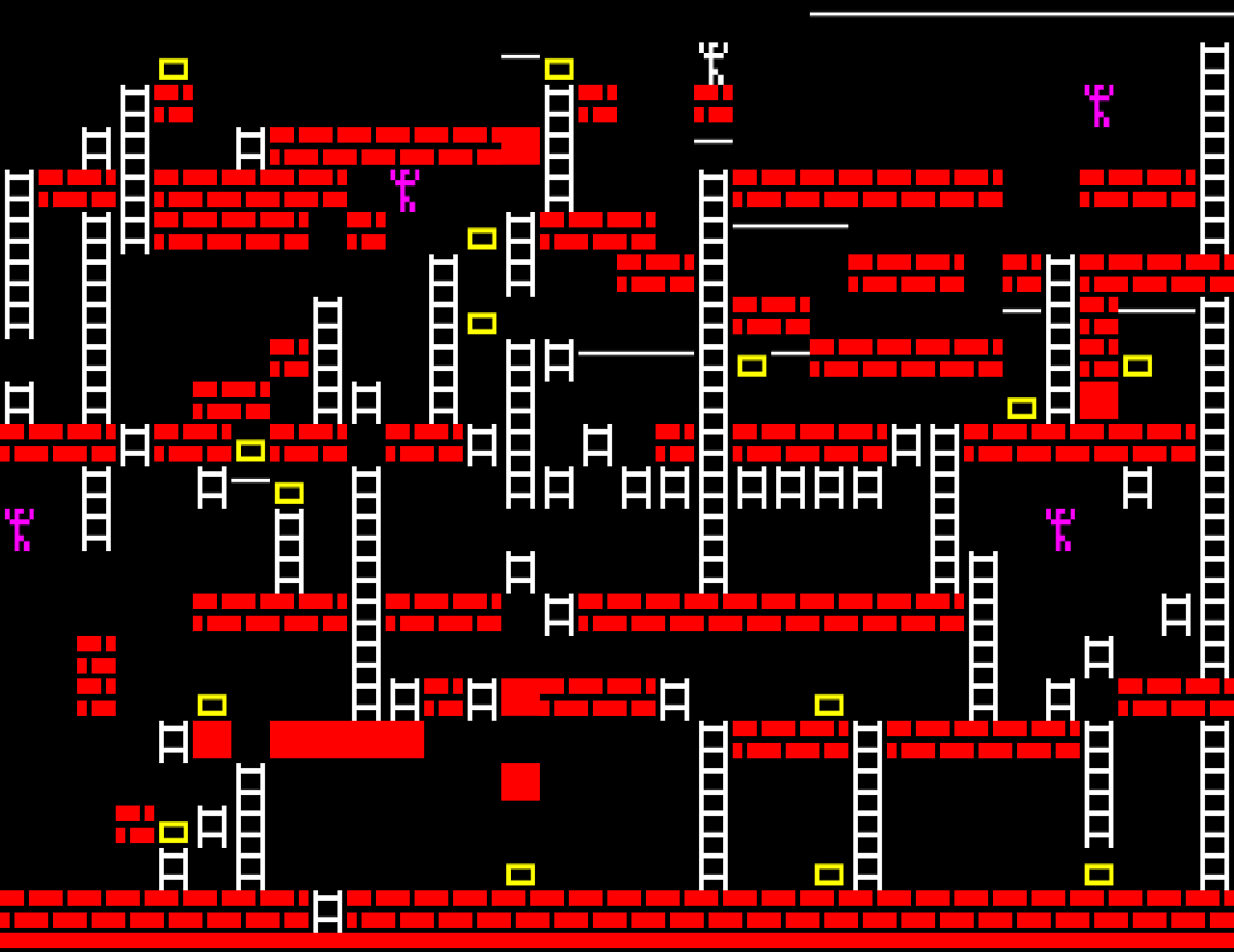}
        \includegraphics[width=.24\linewidth]{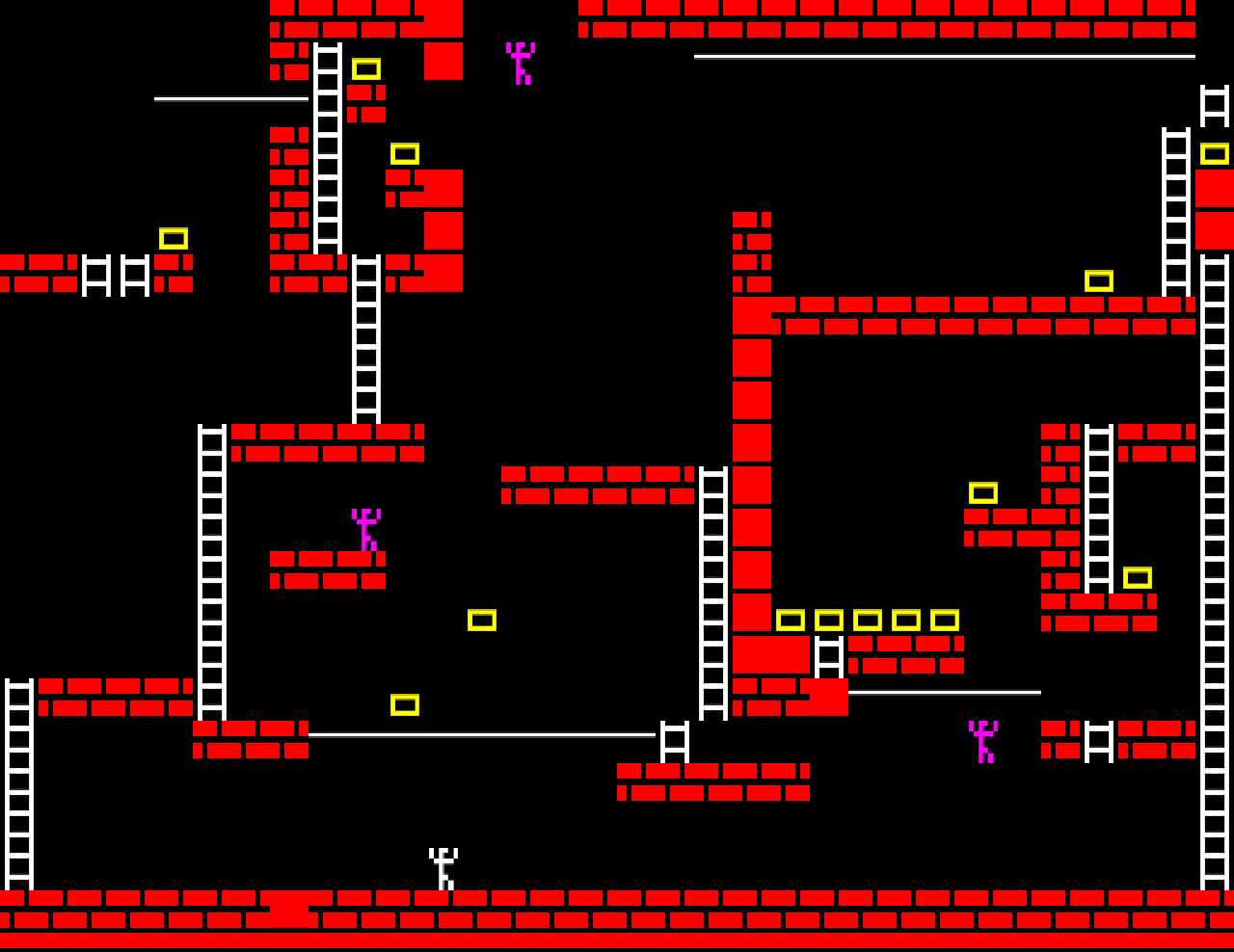}
        \includegraphics[width=.24\linewidth]{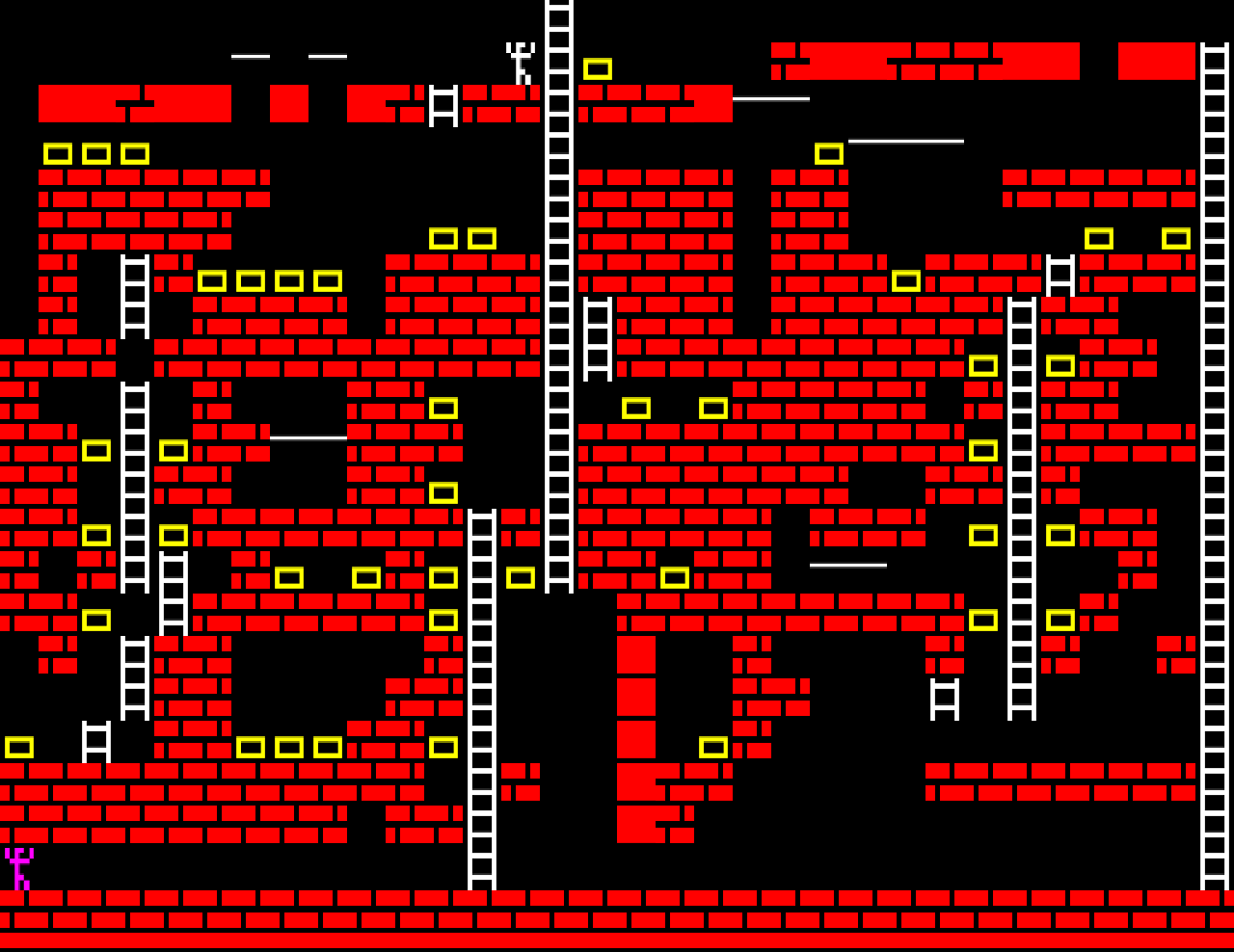}
        \includegraphics[width=.24\linewidth]{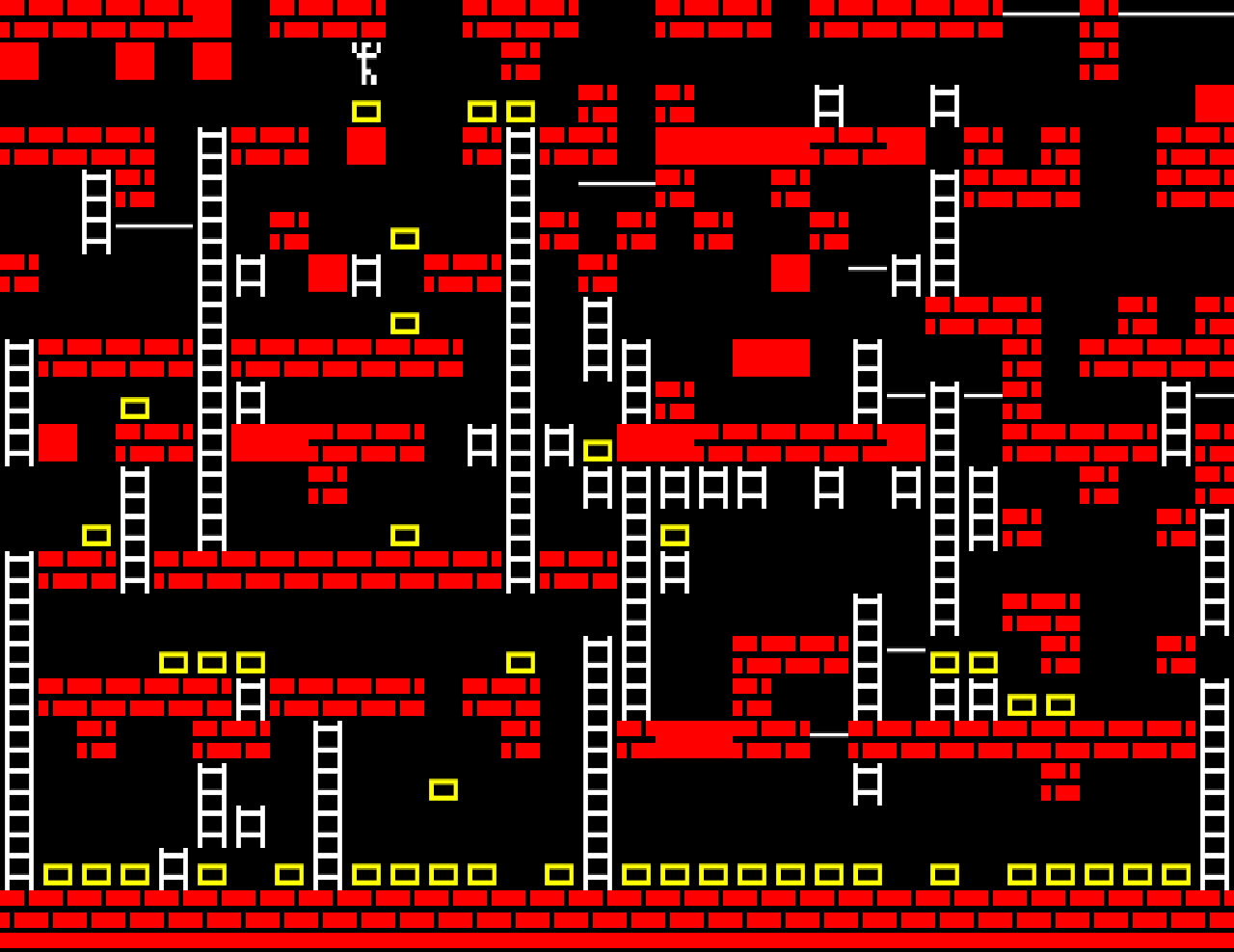}
        \caption{Top four playable levels with respect to originality score. The levels have score of 46.73\%, 48.86\%, 59.66\%, and 60.23\% respectively from left to right.}
        \label{fig:userHigh}
    \end{subfigure}\vspace{.15cm}
    \begin{subfigure}[t]{\linewidth}
        \centering
        \includegraphics[width=.24\linewidth]{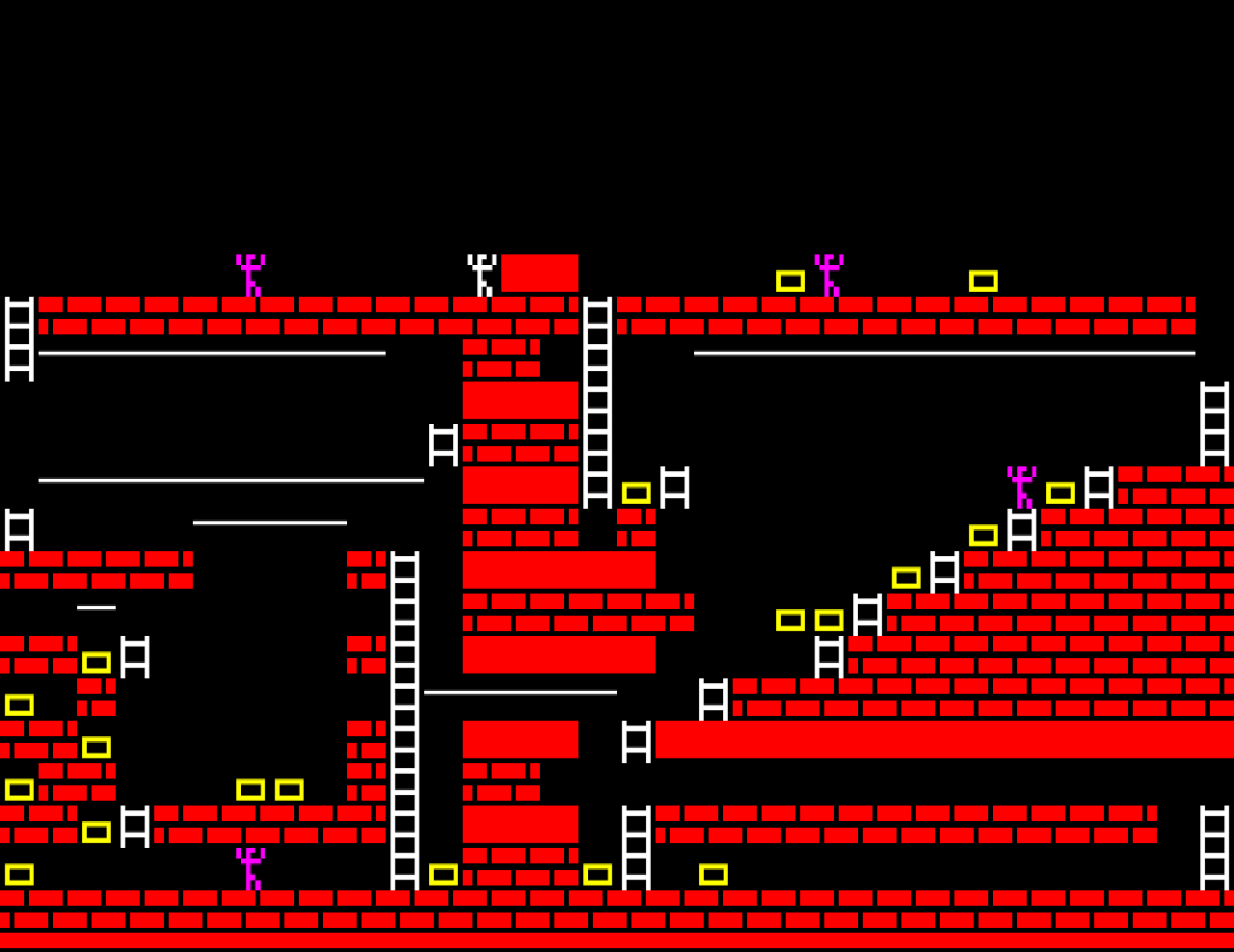}
        \includegraphics[width=.24\linewidth]{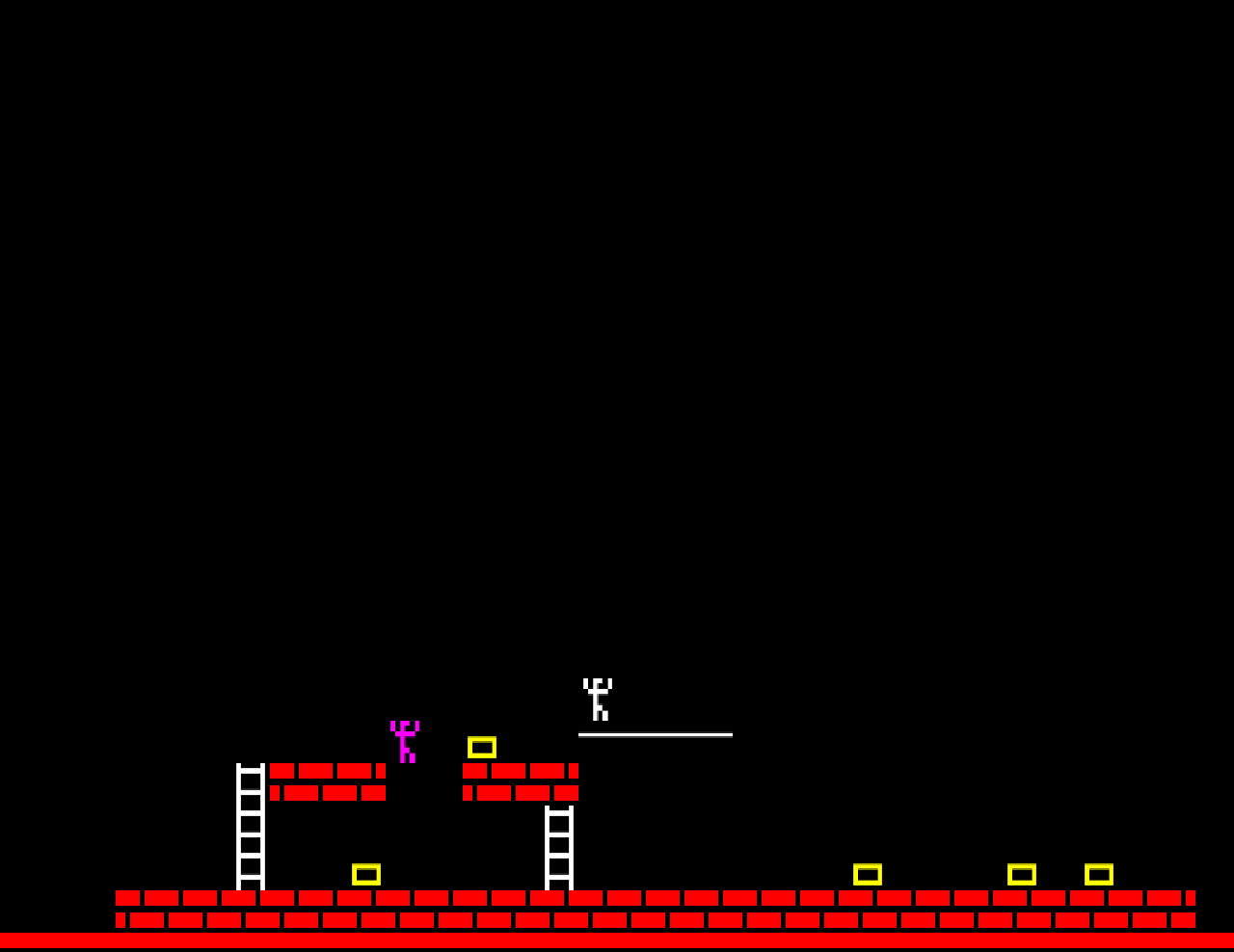}
        \includegraphics[width=.24\linewidth]{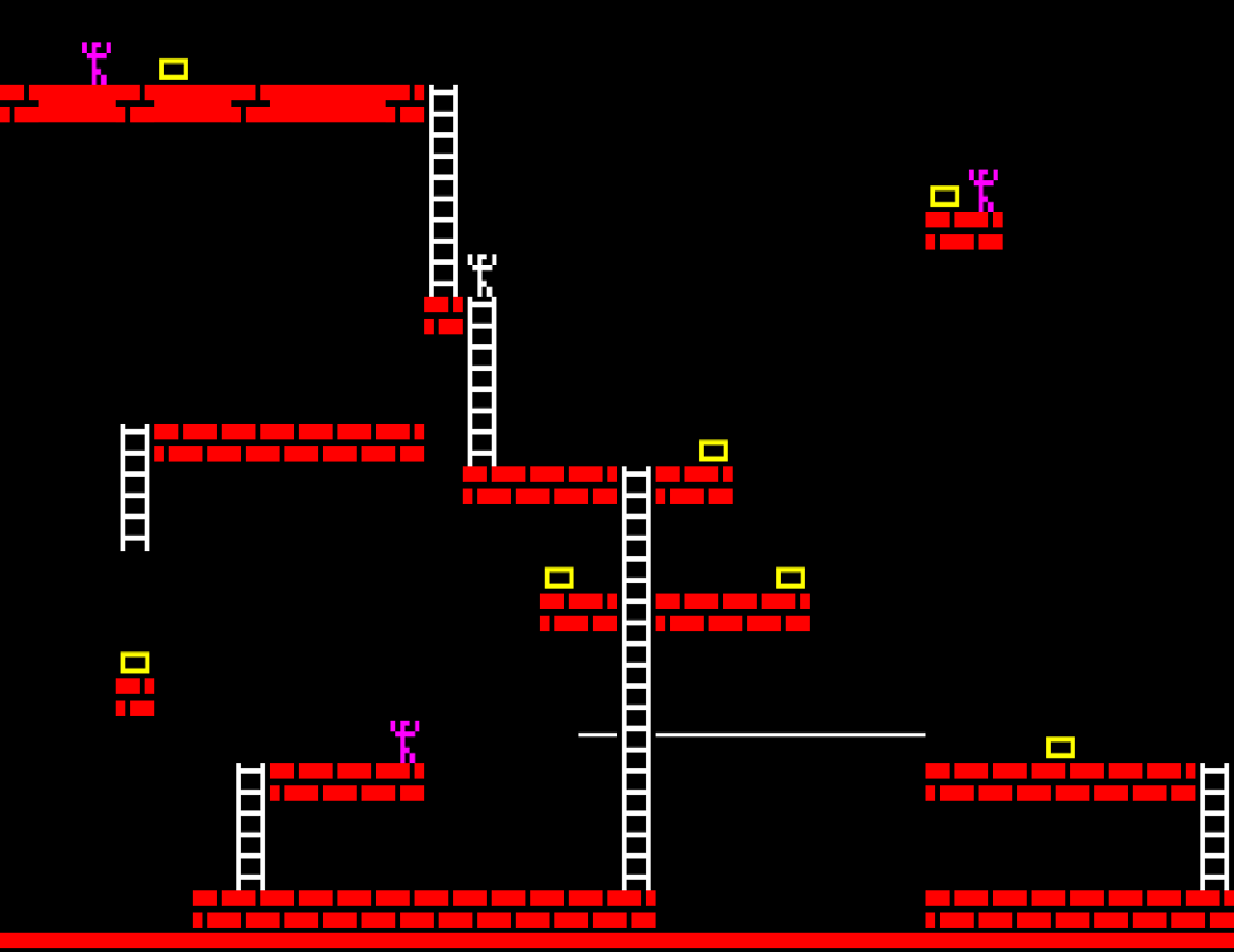}
        \includegraphics[width=.24\linewidth]{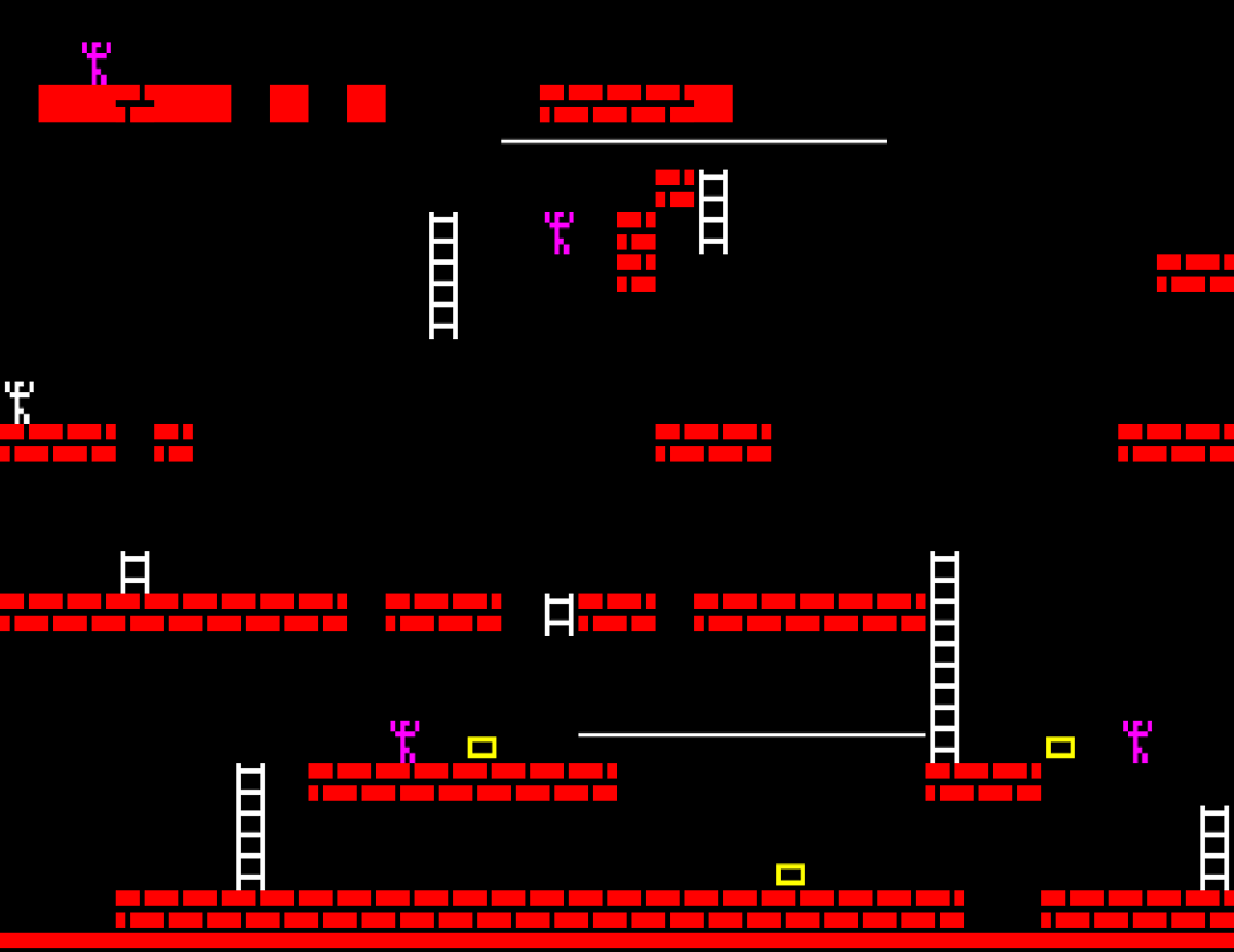}
        \caption{Bottom four playable levels with respect to originality score. The levels have score of 6.96\%, 9.38\%, 17.76\%, and 18.47\% respectively from left to right.}
        \label{fig:userLow}
    \end{subfigure} 
    \caption{Examples of original Lode Runner levels and user created levels}
    \label{fig:userLevels}
\end{figure*}

For our experiment, we released the tool and shared it on Twitter and other social media. We asked user to make their own level, play it and share the level with others. We recorded all the user interaction anonymously in an online database and we also encouraged them to give us their feedback on the tool via an anonymous form (using Google forms). After 18 days from the launch of the tool, we froze the data to analyze it. Table~\ref{tab:stats} show some basic statistics about the collected data. We notice that a lot of people tried the tool but never managed to create playable levels. Looking into the feedback submitted through the feedback form, we think that the main problem for that small conversion ratio is the problem of filling the holes in the design. Many users got frustrated by not being able to fill certain holes in their design because they never occur in the suggestions. We addressed this problem in the next iteration of Lode Encoder by introducing the wand tool. We would love to test the effectiveness of the wand tool on the number of finished playable levels in future research.

Looking into the collected data, we wanted to know which suggestion engine was most frequently used. Figure~\ref{fig:modelUsed} shows the number of interactions that each suggestion got over all the sessions. As we can see, the low variance platform model is the most frequently used by a wide margin. We believe that users preferred this suggestion engine because it has the least changes from their vision due to the low variance part. Also, they preferred this suggestion because it tends to produce more structured levels compared to the rest of the models. It's worth noting that the second most frequently used type of suggestion is the high variance platform model.

As the suggestions from autoencoder is an important part of our system, we tried to see how helpful the suggestions are and how often user try to have new suggestions. Figure~\ref{fig:refreshes} shows a histogram about the number of refreshes used in every session. More than 50\% of the recorded sessions go with the initial suggestions only. Nearly 15\% and 10\% sessions asked for new suggestions once and twice. Few sessions refreshed suggestions more than twice. This histogram also shows that having a limited number of refreshes didn't confine the user during their design process as they usually don't need more than 3 refreshes to complete their level.


We looked into the 24 playable levels, we noticed that only 5 levels have originality score less than 20\% as shown in figure~\ref{fig:originality}. It seems that showing the originality score encouraged users to make levels that are different from the training dataset. Figure~\ref{fig:userLevels} shows the top and the bottom four original levels created by the users. We noticed that the top levels are usually more dense with less empty spaces, while the bottom is the opposite. The originality score might be the culprit behind this as it pushes the users to create levels that is different from the original dataset which is mostly more sparse and full of empty tiles.

\begin{figure*}
    \centering
    \begin{subfigure}[t]{.24\linewidth}
        \centering
        \includegraphics[width=\linewidth]{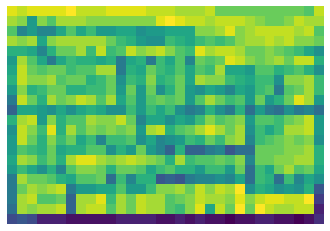}
        \caption{Heatmap of all the empty locations}
        \label{fig:emptyHeatmap}
    \end{subfigure}
    \begin{subfigure}[t]{.24\linewidth}
        \centering
        \includegraphics[width=\linewidth]{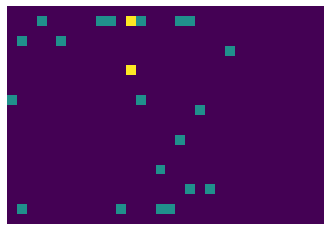}
        \caption{Heatmap of all the player locations}
        \label{fig:playerHeatmap}
    \end{subfigure}
    \begin{subfigure}[t]{.24\linewidth}
        \centering
        \includegraphics[width=\linewidth]{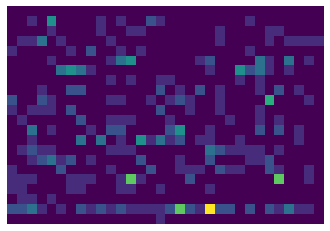}
        \caption{Heatmap of all the gold locations}
        \label{fig:goldHeatmap}
    \end{subfigure}
    \begin{subfigure}[t]{.24\linewidth}
        \centering
        \includegraphics[width=\linewidth]{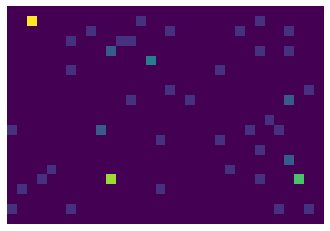}
        \caption{Heatmap of all the enemy locations}
        \label{fig:enemyHeatmap}
    \end{subfigure}
    \caption{Heatmap of the empty locations, the player starting location, gold locations, and enemy locations over the 24 playable levels}
    \label{fig:heatmaps}
\end{figure*}

Finally, we wanted to see if our system was biasing the users toward certain level structures. We decided to layer all the empty tiles in the levels on a heatmap and check if there is any visible structure. Figure~\ref{fig:emptyHeatmap} shows the heatmap. We can notice that there is some bias around always having a ground floor and having similar structures as the  classical lode runner levels. We attribute this to the fact that they look very clean and they are mostly generated by the VAE-platform which was the highest used autoencoder model. We also overlayed the player starting location on a heatmap as shown in figure~\ref{fig:playerHeatmap}. Most of the player starting locations are around the center of the x-axis of the level. We don't know any reason behind this choice but we think that it gives the player equal choice when they start playing compared to starting on the left or starting on the right. Finally, we looked into the gold nuggets locations (in figure~\ref{fig:goldHeatmap}) and the enemy locations (in figure~\ref{fig:enemyHeatmap}), we notice that both are wide spread all over the levels. The only difference is the gold nuggets are more well distributed than enemies, this goes back to the fact there is more gold in a level than enemies.

\section{Conclusion}

This paper presented Lode Encoder, a novel take on AI-assisted mixed-initiative game level design tool. While not the first game design tool to be based on PCGML, it is as far as we can tell the first to attempt the leverage the power of self-supervised learning through autoencoders for assisting game designers. More importantly, however, it proposes an uncommon interaction paradigm. The suggestions are there to empower and inspire the designer, but at the same time the designer is constrained to \textit{only} paint from the suggestions (and the magic wand). 

At the end of the paper, a question in the reader's mind is probably whether this was a successful tool, as this is often what the user study sets out to test. It is safe to say that the tool in its current form is not going to replace existing level editors for various games. However, the formal and informal feedback we have received strongly suggest that many who have tried the tool appreciate its uncommon and game-like mode of interaction. One well-known indie game designer remarked that trying to make a completable level was ``was an interesting and enjoyable task'' and that Lode Encoder is ``more like a game than a tool''.

From another perspective, the paper tells the story of how we ourselves have been trying to design around limitations, namely the limitations of the autoencoder in generating complete and playable levels. We do not know whether we could make autoencoders or some other form of self-supervised learning (e.g. GANs) learn to produce better levels based on the limited number of available Lode Runner levels. However, the Lode Encoder interaction mode is in itself a response to trying to find a use for the imperfect output of our generative models. For this application, it turns out that the slightly and controllably broken levels, that in various ways warp the partial level the user has already constructed, are perfect. We hope that the interaction mode proposed and dissected here can in turn inspire others.

\bibliographystyle{IEEEtran}
\bibliography{IEEEexample}

\end{document}